# Exploring the Role of Artificial Intelligence and Machine Learning in Process Optimization for Chemical Industry

**Zishuo Lin [1], Jiajie Wang [2],Zhe Yan [3]Peiyong Ma [4]**

*Abstract*— **The crucial field of Optical Chemical Structure Recognition (OCSR) aims to transform chemical structure photographs into machine-readable formats so that chemical databases may be efficiently stored and queried. Although a number of OCSR technologies have been created, little is known about how well they work in different picture deterioration scenarios. In this work, a new dataset of chemically structured images that have been systematically harmed graphically by compression, noise, distortion, and black overlays is presented. On these subsets, publicly accessible OCSR tools were thoroughly tested to determine how resilient they were to unfavorable circumstances. The outcomes show notable performance variation, underscoring each tool's advantages and disadvantages. Interestingly, MolScribe performed best under heavy compression (55.8% at 99%) and had the highest identification rate on undamaged photos (94.6%). MolVec performed exceptionally well against noise and black overlay (86.8% at 40%), although it declined under extreme distortion (<70%). With recognition rates below 30%, Decimer demonstrated strong sensitivity to noise and black overlay, but Imago had the lowest baseline accuracy (73.6%). The creative assessment of this study offers important new information about how well the OCSR tool performs when images deteriorate, as well as useful standards for tool development in the future.**

## I. Introduction

Chemical journals contain a great deal of useful information that predates the common practice of annotation and curation. Literature databases are now essential for storing and retrieving chemical information due to the exponential increase of scientific literature and the rise in the number of chemical publications. Chemical data is frequently represented graphically as chemical structure images, which are easily interpreted by humans. However, in order to facilitate effective database querying and retrieval, these graphical representations must be transformed into machine-readable formats called chemical structure identifiers. Optical Chemical Structure Recognition (OCSR) is an area that has witnessed the development of many instruments and techniques to address this difficulty.

Researchers at UCT Prague's Department of Informatics and Chemistry are working on a noteworthy project in this area: creating a database of chemical structures created by Czech scientists. Accurately annotating chemical structures in publications using OCSR is a crucial component of this endeavor. Nevertheless, a large number of these publications are only available as scanned physical papers, which might result in graphical degradation like noise, distortion, or compression artifacts. These deteriorations have a significant effect on OCSR instruments' accuracy, hence a thorough assessment of how well they function in these circumstances is required.

By concentrating on how well open-source OCSR tools operate under various levels of graphical damage, this study fills a significant research vacuum. This was accomplished by developing a new testing dataset that included chemically structured photos with gradually added artificial damage to mimic real-world situations. Finding the most reliable OCSR tools for chemical structure recognition, measuring the effects of different picture degradations, and offering useful information for choosing the best tools for real-world applications are among the goals.

This research explores the resilience of OCSR tools under increasingly difficult settings, providing a thorough empirical comparison, in contrast to other studies that mostly assessed tools on datasets that were undamaged or just slightly affected. The following is the format of the following chapters: The theoretical basis, including the fundamental ideas of OCSR and its applicability, is given in Section II. In Section III, relevant literature is reviewed, emphasizing significant developments and current constraints in OCSR research. The materials and techniques used, such as the development of the testing dataset and the configuration of the OCSR tools, are described in Section IV. The experimental analysis, which displays the performance outcomes and insights, is presented in Section V. Section VI ends with a review of the results, recommendations for choosing OCSR tools, and future research topics, including possible ways to enhance OCSR.

## II. Theoretical Background

The inclusion of chemical structures in publications serves a critical purpose: to convey information that is clear, accurate, and comprehensible to readers. While guidelines exist for creating these representations, the sheer flexibility in how a single molecule can be depicted—such as the many valid ways to draw benzene (refer to Fig. 1)—means no single format can achieve universal standardization. In most chemical publications, typically available in Portable Document Format (PDF), chemical structures are presented either as raster or vector images, with raster formats being more prevalent. These images are either directly exported from specialized editing software or generated from scans of physical prints, ensuring accessibility across various mediums and formats.

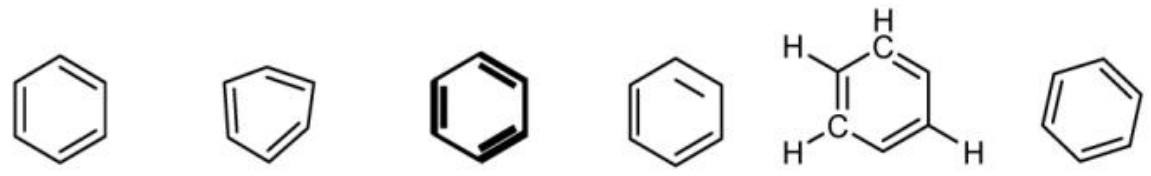

Fig. 1. Six legitimate benzene representations

However, the purpose of vector and raster representations of chemical structures is to ensure their identification by humans. In contrast, chemical structure identifiers are designed to be recognized and processed by computers, enabling effective storage and querying within chemical databases. To achieve this, these identifiers must be standardized, easily storable, and convertible across various formats, including the aforementioned raster and vector images. The methods employed to fulfill these requirements can generally be categorized into line notations and connection tables. Line notations serve as concise, single-string representations of chemical structures and are fundamental in the field of cheminformatics. The most widely used systems for this purpose are the International Chemical Identifier (InChI) and the Simplified Molecular Input Line Entry System (SMILES), along with its numerous derivatives. SMILES is a highly effective system for encoding two-dimensional chemical structures by constructing a connection graph of the molecular structure. It is intentionally designed to be intuitive for humans to understand and straightforward for computer algorithms to generate. For instance, ethanol is represented simply as "CCO", requiring only three bytes of data. This simplicity has solidified SMILES as the most widely used method for encoding chemical structures to date. However, its flexibility comes with a drawback: the same structure can be represented in multiple ways. For instance, ethanol can also be written as "OCC" or "CCO", meaning that without standardization, two independent algorithms may only match SMILES strings 50% of the time for smaller molecules, with accuracy decreasing as molecular complexity increases. To address this limitation, derivatives of SMILES such as SMARTS, DeepSMILES, and SELFIES have been developed, offering a 1:1 structure-to-representation relationship and greater consistency. Whereas, InChI employs a hierarchical, layered approach to encoding chemical structures. Each structure is represented as a single-line string, organized into distinct layers, with each layer prefixed by a forward slash. For instance, in the case of guanine (refer to Fig. 2), the InChI string begins with "InChI" to indicate the identifier standard used, followed by layers for molecular formula, connectivity, isotopes, stereochemistry, and tautomers. Unlike SMILES, InChI is specifically designed to be machine-readable rather than human-friendly. Its key advantage lies in its precision: a given molecule has exactly one unique InChI representation. This consistency makes InChI particularly powerful for tasks like database searching and ensuring unambiguous identification of chemical structures.

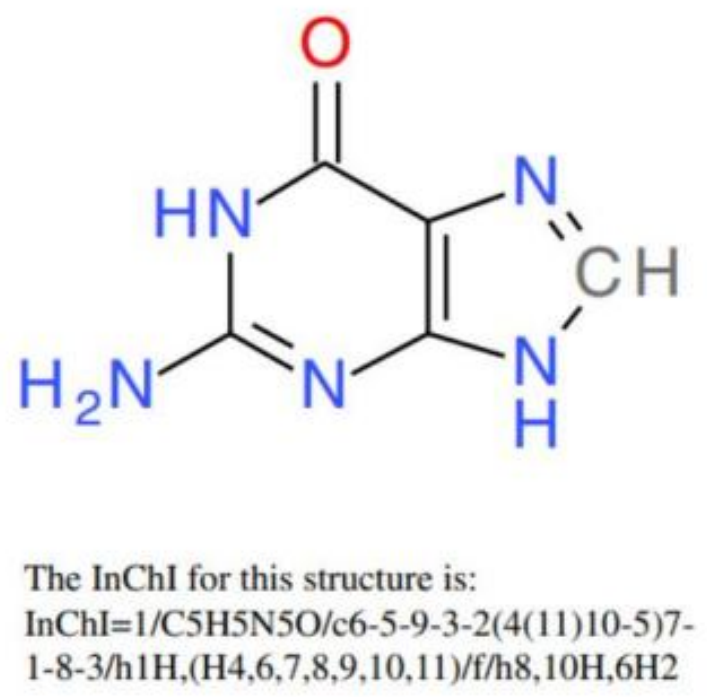

Fig. 2. Guanine's chemical makeup and matching InChI encoding

The connection table approach defines chemical structures by representing atoms with three-dimensional coordinates ($x$, $y$, $z$) and their mutual connectivity. This method closely aligns with graphical representations, making the generation of visual molecular structures from connection tables straightforward and efficient. While various implementations of this approach exist, MDL molfiles have emerged as the most widely adopted standard, underscoring their reliability and effectiveness in encoding and visualizing molecular structures. A molfile is structured into three distinct sections. First, the header block includes the title, timestamp, and optional comments. Second, the connection table represents the actual molecular structure, and finally, the file ends with the line "M END". This structure is best illustrated through examples, such as the MDL molfile V2000 representation of leucine[1] generated by ChemDraw[2] (refer to Fig. 3). While molfiles are widely used, they do have limitations. One key drawback is the simplified depiction of chemical bonds, which are restricted to single, double, and triple covalent bonds. Additionally, challenges arise in handling implicitly stated hydrogen atoms, particularly when combined with non-trivial valency, potentially leading to misinterpretation of the implied hydrogens. Despite these limitations, molfiles remain a standard format in cheminformatics due to their versatility and widespread adoption.

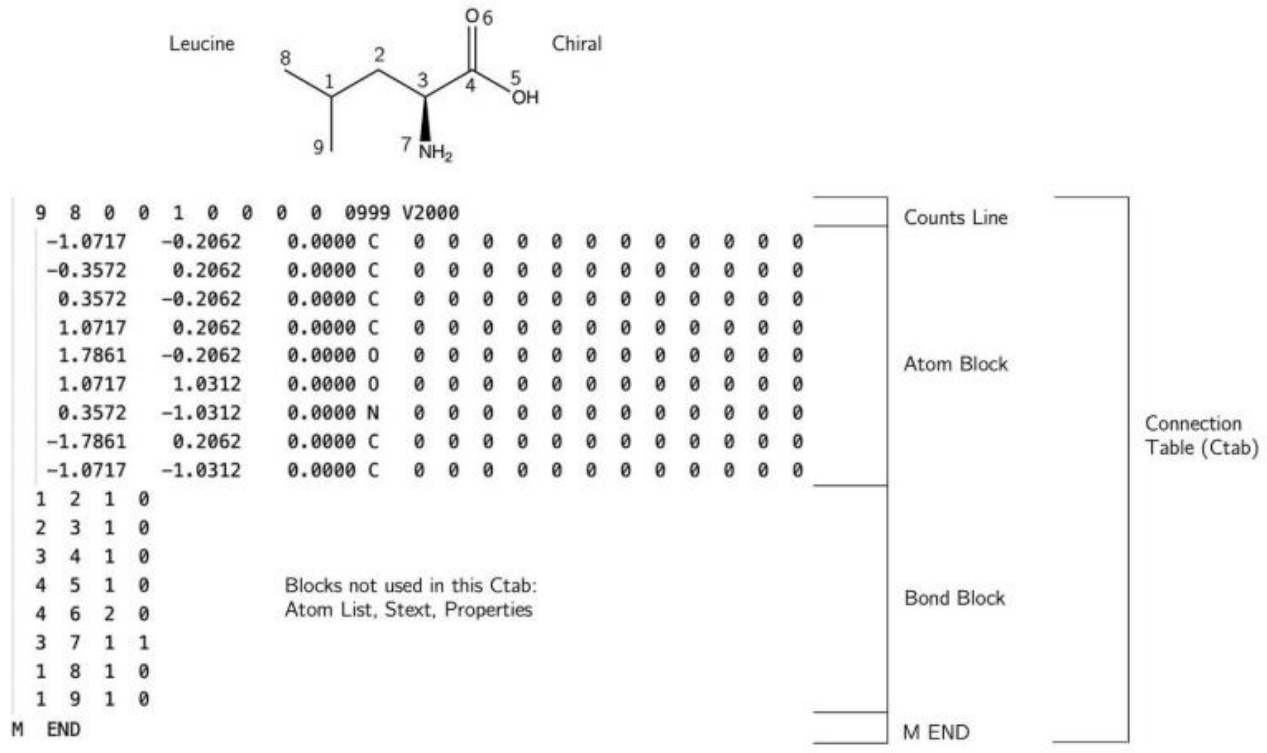

Fig. 3. An illustration of a connection table and end line for the molecule leucine produced by ChemDraw in an MDL molfile V2000

[1] https://en.wikipedia.org/wiki/Leucine
[2] https://revvitysignals.com/products/research/chemdraw

## III. Related Works

OCSR serves as a critical intermediary between human-readable formats and machine-readable chemical structure identifiers such as [1]. This process typically involves three key steps [2]: identifying chemical structures within a document while distinguishing them from other graphical elements, compiling these structures into chemical graphs, and interpreting them into standard chemical structure identifiers. The first two steps fall under segmentation, while the final step focuses on recognition and output generation. OCSR tools are designed to either handle both segmentation [3] and recognition [4] or specialize exclusively in the recognition process. In the latter case, the input must already be preprocessed into a single image of the chemical structure. Therefore, this study concentrates specifically on the recognition phase, which is essential for generating precise and accurate chemical identifiers. The methodologies for achieving the goals of OCSR can be broadly classified into two main categories. Earlier tools predominantly relied on rule-based approaches [5], [6], leveraging predefined algorithms and heuristics to process chemical structure images. In contrast, more recent advancements have embraced Machine Learning (ML)-based systems, applying sophisticated models to tackle the problem of image captioning for chemical structures. These ML approaches offer greater flexibility and adaptability, marking a significant evolution in OCSR technology such as [7], [8].

One of the pioneering systems to address the needs of OCSR was [9], introduced in 1992. Its workflow set the foundation for many subsequent rule-based systems. [9] process involves scanning the input image, vectorizing it through raster-to-vector conversion, detecting bond lines, applying OCSR to identify atoms, constructing a chemical graph based on the gathered data, and generating outputs in multiple machine-readable formats, such as SMILES. This step-by-step methodology has become a standard framework for rule-based OCSR systems, with variations or enhancements distinguishing individual tools. As illustrated in Fig. 4, taken from a 2020 review of OCSR tools, most early systems were commercial. Therefore, this study, however, focuses exclusively on freely available tools. The Optical Structure Recognition Application[3] (OSRA) was the first open-source tool developed for OCSR. Its workflow follows the general rule-based structure, and its open-source nature has significantly contributed to the development of both OSRA and the broader OCSR field. OSRA is versatile in its input handling, requiring no specific image specifications such as resolution, color depth, or font type. Leveraging the ImageMagick[4] library, it can process a wide range of formats, including TIFF, JPEG, GIF, PNG, Postscript, and PDF. OSRA is primarily implemented as a command-line utility, with a web interface available for demonstration purposes, though it remains under development. The most recent version, as of 2023, is 2.1.0. However, a significant challenge in using OSRA, compared to other OCSR tools, is its complex compilation process. It requires the installation of several version-specific dependencies (including GraphicsMagick[5], POTRACE[6], GOCR[7], TCLAP[8], and OpenBabel[9]), all of which must be compiled from source code. This issue has been highlighted in various benchmarking and implementation attempts. For instance, a 2020 OCSR review benchmarked OSRA using a PyOSRA[10] environment, and a conda recipe for OSRA 2.1.0 is available through bioconda. However, despite these resources, none of the installation methods outlined for this study have been successful.

| Tool name | Programming language used | Operating System compatibility | Open-source | Commercial or free availability (2020) | Ongoing development |
|---|---|---|---|---|---|
| Kekulé | C++ | Windows | No | Yes | No |
| OROCS | C | IBM OS/2 | No | No | No |
| CLiDE Pro | C++ | Windows | No | Yes | Yes |
| OSRA | C++ | Independent | Yes[a] | Yes | Yes |
| ChemReader | C++ | Windows | No | No | No |
| MolRec | Unknown | Unknown | No | No | Unknown |
| Imago | C++ | Independent | Yes | Yes | No |
| ChemOCR | Java | Independent | No | Yes | Yes |
| ChemInfty | Unknown | Windows | No | No | No |
| eChem | Unknown | Unknown | No | No | No |
| MLOCSR | Unknown | Only Web interface | No | Only web interface | Unknown |
| OCSR | Unknown | Unknown | No | No | Unknown |
| ChemRobot | Unknown | Unknown | No | No | Unknown |
| MolVec | Java | Independent | Yes | Yes | Yes |
| MSE-DUDL | Python | Independent | No | No | No |
| Chemgrapher | Python | Independent | No | No | Yes |

Fig. 4. A comparison of the instruments and techniques used in the 2020 OCSR Review

The release of OSRA significantly facilitated the development of subsequent OCSR systems, one of which is Imago[11]—an open-source toolkit for 2D chemical structure image recognition. Imago was designed with the goal of creating a cross-platform library suitable for various applications, including those on mobile devices. Unlike OSRA, Imago has no external dependencies, which is reflected in its straightforward installation process, requiring only the download of a single executable file. Its workflow follows the familiar pattern of other rule-based systems and offers flexibility, allowing it to be used both as a command-line utility and as a Graphical User Interface (GUI) program. The most recent version of Imago is 2.0.0, though the related publication does not mention plans for further development. One of the key focuses of Imago, relevant to this study, is the impact of noise and graphical damage on recognition accuracy. We emphasize the detrimental effects of low resolution, a limited number of symbols, and atom labels containing multiple symbols, all of which significantly reduce correct recognitions. These challenges have become a central concern in the ongoing development of OCSR tools, as improving recognition under such conditions is critical for advancing the technology. In 2019, the MolVec[12] Java library was developed to meet the demand for a lightweight, fully self-contained, and accessible OCSR tool that does not require advanced programming knowledge for implementation. MolVec is solely a recognition tool, meaning it lacks a segmentation module and can only process images contain a single chemical structure. In addition to its use as a Java library, MolVec provides a command-line runnable Main class. However, development appears to have ceased, with the latest release, version 0.9.8, issued on October 14, 2020. Despite this, MolVec remains a valuable tool for straightforward chemical structure recognition.

[3] https://sourceforge.net/p/osra/wiki/Home/
[4] https://imagemagick.org/script/index.php
[5] http://www.graphicsmagick.org/
[6] https://potrace.sourceforge.net/
[7] https://jocr.sourceforge.net/
[8] https://tclap.sourceforge.net/
[9] https://openbabel.org/index.html
[10] https://github.com/edbeard/pyosra
[11] https://www.imago-images.com/
[12] https://github.com/ncats/molvec

Rule-based systems operate by applying a set of predefined rules to make decisions, yielding reliable results when handling simpler datasets with structures that can be easily described by a limited number of rules. However, as chemical structures grow in complexity, predicting unique sub-structures or handling rule exceptions becomes increasingly difficult. The inherent variation in chemical structures found in publications presents a significant challenge for rule-based systems, making it impractical to account for all possibilities using only predefined rules. This is where ML-based OCSR steps in, approaching the problem as an image captioning task by leveraging neural networks—a data-driven, Deep Learning (DL) methodology. A widely used technique in image captioning is the encoder-decoder network, where a Convolutional Neural Network (CNN) [10], [11], [12] serves as the encoder to extract image features, while a Recurrent Neural Network (RNN) [13], [14] acts as the decoder to interpret these features and generate textual output. Alternatives, such as Long Short-Term Memory (LSTM) [15] or Gated Recurrent Unit (GRU) [16] networks, may replace the RNN for certain tasks. Moreover, image features are weighted through attention mechanisms, allowing the model to focus on the most significant aspects of the image. The Transformer [17] model, which incorporates a self-attention mechanism, has demonstrated exceptional performance and is rapidly becoming the foundation for advanced ML-based OCSR systems. For instance, the DECIMER[13] project continues to evolve, with plans to integrate advanced DL technologies and expand its training datasets to 50-100 million chemical structures. They anticipate that this significant growth in data and technology will further enhance its performance. MolMiner[14] utilizes neural networks designed for semantic segmentation and object detection tasks, which are adapted for the recognition of atom and bond elements within chemical documents. These recognized elements are subsequently assembled into a molecular graph through a distance-based construction algorithm. SwinOCSR[15], an open-source ML-based system leverages the Swin Transformer as the backbone of its model, followed by a Transformer encoder and decoder. The Swin Transformer extracts image features into a high-dimensional patch sequence, which is then flattened and fed into the Transformer encoder. The decoder subsequently generates the corresponding DeepSMILES[16]. One of the key advantages of this approach is that, instead of using pooling—commonly employed in CNNs and prone to information loss—the Swin Transformer merges neighboring patches of the sequence, reducing the size of feature maps and preventing the loss of crucial information.

[13] https://decimer.ai/
[14] https://github.com/gorgitko/molminer
[15] https://github.com/suanfaxiaohuo/SwinOCSR
[16] https://github.com/baoilleach/deepsmiles

## IV. Materials And Methods

### A. Dataset Analysis

To evaluate the selected OCSR tools, a dataset of chemical structures was curated. The objective was to compile a representative set of chemical structure images and subject them to various types and degrees of graphical damage. This approach aimed to determine which forms of damage would adversely impact the recognition capabilities of each OCSR tool. The first step in creating the dataset involved selecting the chemical structures to be included. The molecules were chosen using ChatGPT[17], employing a combination of prompting and trial and error to generate a diverse set of trivial or systemic names. A significant number of suggestions were produced, and a carefully curated selection was made, ensuring variation in substituents, chain lengths, and bond configurations. The final dataset was composed of 129 structures, organized as follows: 20 simple linear molecules, 20 simple branched molecules, and 30 cyclic molecules—further subdivided into categories such as simple structures with a benzene core, heterocyclic structures, fused rings, bridged rings, polycyclic structures, and macrocyclic structures. Additionally, the dataset included 40 biochemically relevant molecules, encompassing amino acids, saccharides, lipids, hormones, vitamins, and unique or unusual structures. Ten molecules combining linear, branched, and cyclic elements, along with nine pharmaceutically significant molecules, were also incorporated. This selection of 129 systematic or trivial names formed the foundation of the dataset, captured in a .txt file for further processing.

The selected chemical names generated by ChatGPT were utilized to create their corresponding chemical structures using the "Convert Name to Structure" command in ChemDraw Professional version 22.2.0. ChemDraw was then employed to export images of the structures as 300 dpi TIFF files. The dimensions, size, and resolution of the images were not explicitly specified, as ChemDraw automatically determined these parameters to ensure consistent padding around the outer edges of the molecules. Additionally, ChemDraw was used to export the corresponding molfiles, which would serve as reference data for subsequent testing. This process resulted in the creation of a base dataset comprising 129 TIFF files and their corresponding molfiles, providing a foundation for further analysis and evaluation. Compression is one of the most prevalent forms of graphical degradation that stored images may experience, with JPEG compression being the most commonly used method. To simulate this type of damage, the Image module from the Pillow library[18] (a fork of PIL) was utilized. The original TIFF images were processed using the Image.save() function, with the quality parameter adjusted to values of 80, 60, 40, 20, and 1. This process generated five subsets of images labeled as tiffs_compressed_(compression percentage). An example of the progressive degradation caused by varying levels of compression is illustrated in Fig. 5, demonstrating the impact on image quality as compression increases.

[17] https://openai.com/index/chatgpt/
[18] https://pillow.readthedocs.io/en/stable/

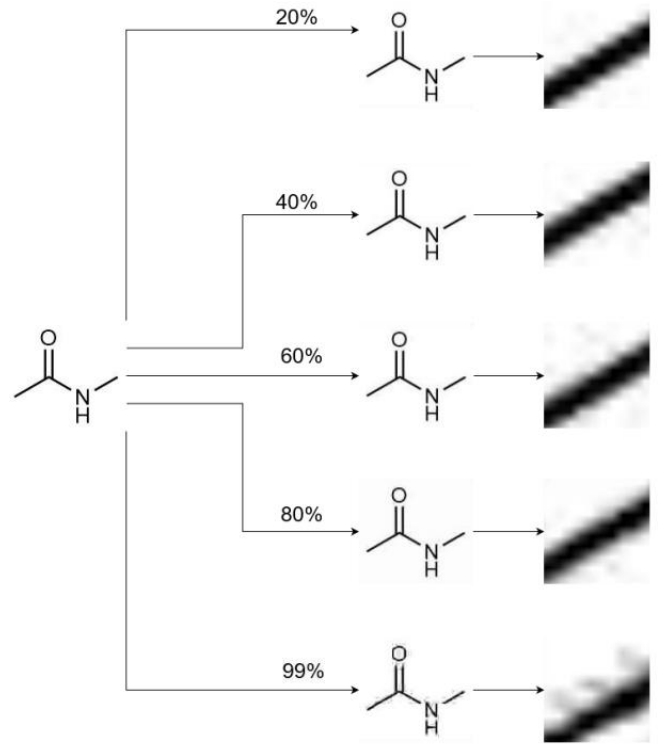

Fig. 5. Illustration of progressive compression damage

To replicate the common issue of scanned files featuring backgrounds that are not perfectly white, a subset of images was created by progressively overlaying a black mask at varying intensities (20%, 40%, 60%, and 80%). This simulation was achieved using the blend function from ImageMagick. The resulting subsets were labeled as tiffs_blend_(overlay_percentage). Refer to Fig. 6 for a visual example of this gradual overlay effect. Imperfect scanning processes often result in creases and distortions within the original image. To simulate such damage, a distortion effect was applied using the img.distort() function from ImageMagick, with the distortion method specified as 'Shepards' and points selected randomly. To ensure the distorted structures remained within the image frame, a 30-pixel white padding was added to all images prior to applying the distortion. Shepard's distortion moves a specified source point to a corresponding destination coordinate. For this simulation, a source point was chosen as a random *x*, *y* coordinate within the range of 0.3 to 0.7 times the maximum *x* or *y* dimensions of the image. The destination point was randomly placed at a distance from the source point, with the scale of the distance increasing according to a scale parameter (ranging from 0.1 to 0.5). This scale parameter was selected arbitrarily based on the resulting quality of the distortions. The outcome of this process was the creation of subsets labeled tiffs_distort_(scale_parameter). See Fig. 7 for an example of the distortions. Notably, an unexpected outcome of this distortion method was the generation of "hand-drawn-like" representations of chemical structures through a quick and automated procedure. Examples of such images are illustrated in Fig. 8.

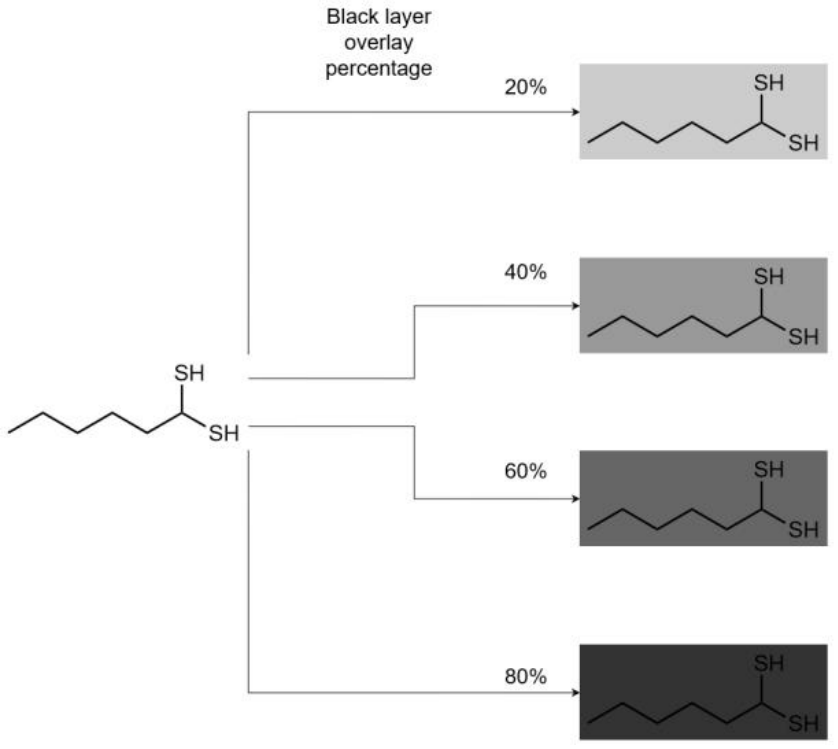

Fig. 6. Gradual deterioration is visualized by blending a black layer

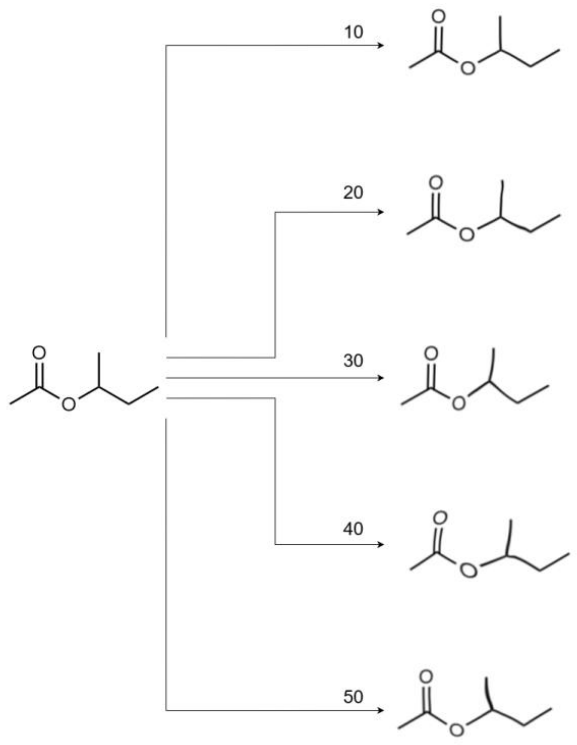

Fig. 7. Shepards distortion is used to visualize gradual distortion

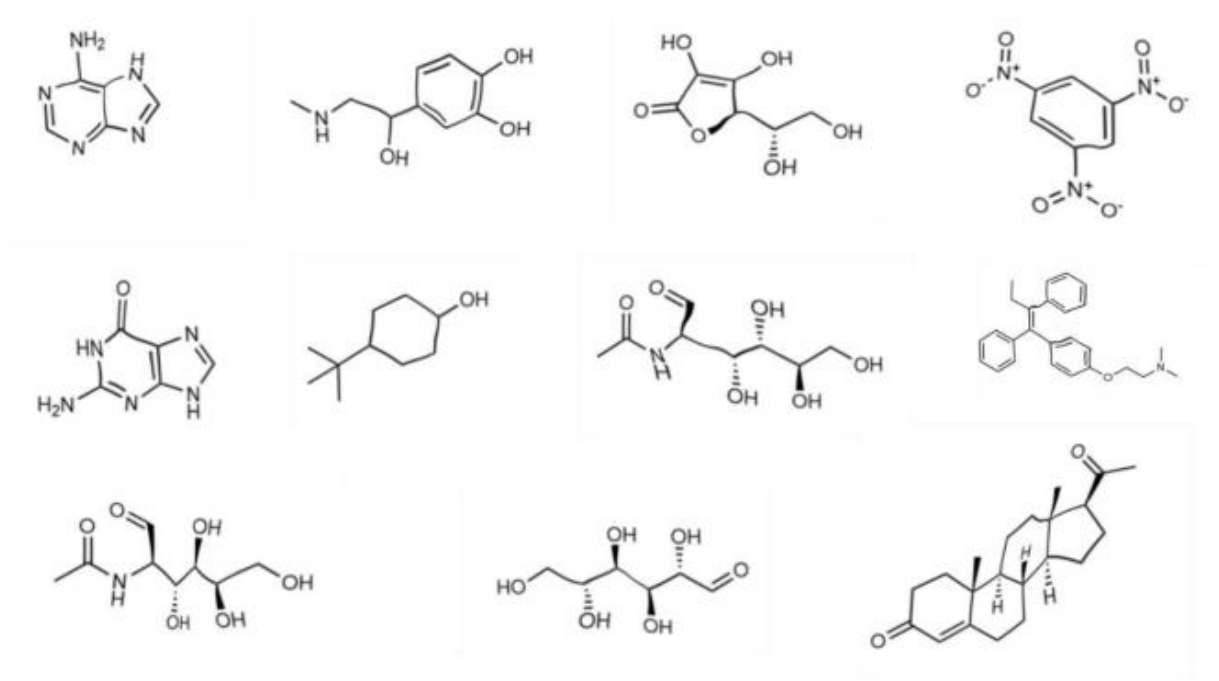

Fig. 8. Examples of warped structures that look like chemical drawings by hand

## B. *Model Analysis*

The primary objective of the testing phase was to evaluate the recognition success rate of each OCSR tool across all dataset subsets. This was accomplished by comparing the molfiles generated by the tools during the recognition process to the reference molfiles exported from ChemDraw. The general workflow for testing can be outlined as follows: TIFF files containing chemical structure images were provided as input to the OCSR tools. Each tool attempted to recognize the structures within the images and generate a corresponding chemical structure identifier, ideally in the form of a molfile. This process was repeated for all subsets of the dataset. In cases where the recognition process failed, was aborted, entered a processing loop, or resulted in an empty molfile, a placeholder molfile was generated to signify unsuccessful recognition for that specific attempt. After processing all images, the generated molfiles from each subset were compared to the reference molfiles to assess recognition accuracy. This workflow is illustrated in Fig. 9.

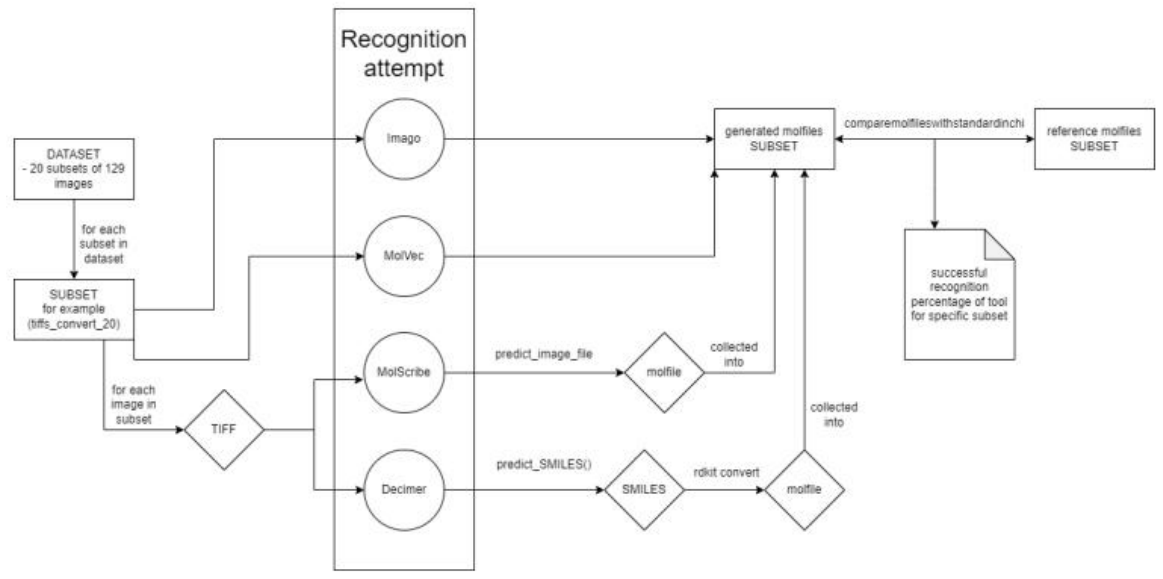

Fig. 9. OCSR tool testing visualization

Each OCSR tool was installed and tested within its own dedicated Miniconda environment to prevent any conflicts arising from overlapping dependencies. Since execution time was assessed on a relative scale rather than being precisely measured, the specific processor and graphics card used were not critical to the testing process. After all recognized molfiles were generated, a modified version of a script from the 2020 OCSR benchmarking study was employed for evaluation. This script processes both the generated and reference molfiles, converting them into their corresponding InChI strings using RDKit[19]. It then assesses whether the two InChI strings match and records the evaluation result, along with both the generated and reference InChI strings, in a text file shared for the entire subset. Once all molfiles within a given subset have been compared, the evaluation outcome is calculated as the ratio of matching InChI strings to the total number of molfiles in the subset. This provides a quantitative measure of the recognition success rate for each tool. Certain tool-specific behaviors required manual intervention to ensure the recognition and evaluation processes were not interrupted. For Imago, the tool entered a processing loop while attempting to recognize three structures within the tiffs_convert_20 subset, though this issue did not occur with the more degraded tiffs_convert_25 subset. To mitigate this, the problematic structures were temporarily removed from the dataset and replaced with duplicates of randomly selected, different structures from the subset. This ensured that the recognition attempt would still be evaluated as unsuccessful without disrupting the overall process. For DECIMER, the only available output format for recognition was SMILES strings generated by the predict_SMILES() function. A custom testing script was developed to facilitate the conversion of SMILES strings into molfiles, ensuring consistency with the evaluation process used for other tools. In cases where DECIMER produced invalid SMILES strings that caused parsing errors during molfile conversion, the script generated a placeholder molfile, resulting in the recognition being evaluated as unsuccessful.

## V. Experimental Analysis

This section primarily examines the impact of various types of graphical damage on the performance of OCSR tools. The analysis is conducted by comparing the recognition success rates achieved on subsets with induced graphical damage to those obtained on the corresponding unmodified subsets. The introduction of even a slight degree of black overlay, compression damage, or noise resulted in a significant decline in performance as shown in Table I and Fig. 10. Higher levels of compression or noise caused recognition rates to drop substantially. Similarly, distortion damage led to a gradual decrease in recognition success. The presence of a black overlay did not appear to affect the recognition success of MolVec as shown in Table II and Fig. 11. Additionally, MolVec demonstrated a notable resilience to compression damage and noise, with only higher levels of such damage resulting in a significant decrease in recognition rate. However, the introduction of any distortion caused an approximate 20% reduction in the success rate, with further increases in distortion leading to progressively lower recognition success.

TABLE I
RECOGNITION RATES PER SUBSET OF IMAGO

| Subset | Base | _20 | _40 | _60 | _80 | _99 | _5 | _10 | _15 | _20 | _25 | _30 |
|---|---|---|---|---|---|---|---|---|---|---|---|---|
| Blend | 73.6% | 34.9% | 33.3% | 32.6% | 29.5% | - | - | - | - | - | - | - |
| Compress | 73.6% | 37.2% | 37.2% | 36.4% | 34.9% | 20.2% | - | - | - | - | - | - |
| Convert | 73.6% | - | - | - | - | - | 34.9% | 38.0% | 15.5% | 6.2% | 9.3% | - |
| Distort | 73.6% | 62.8% | 64.3% | 59.7% | 55.8% | - | - | - | - | - | - | 54.3% |

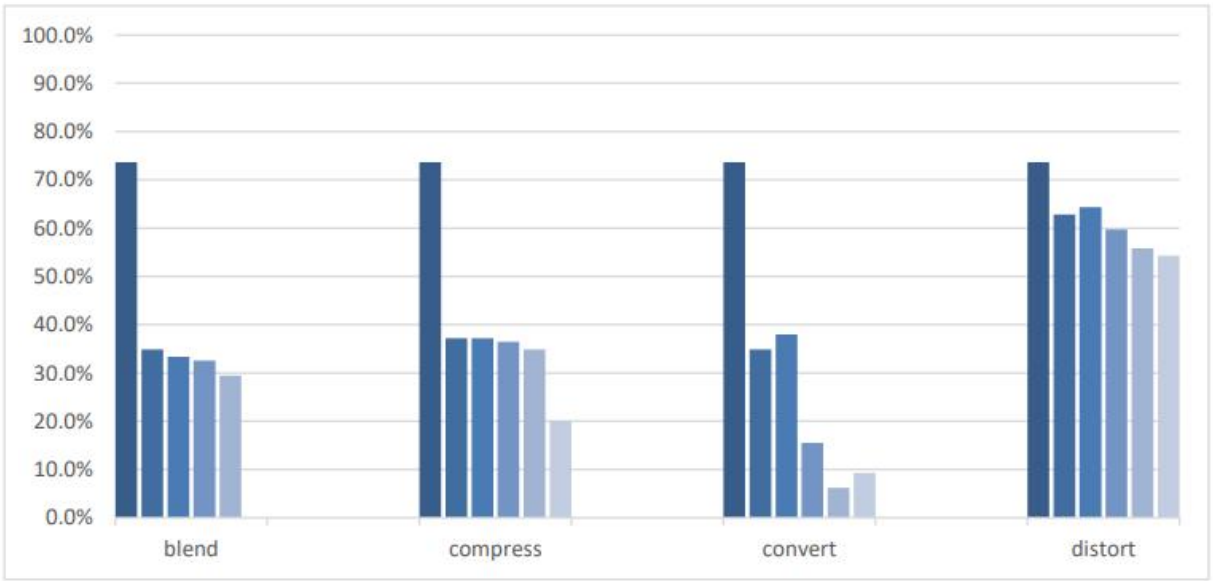

Fig. 10. Test results for Imago subsets impacted by a particular kind of damage (the first column shows a subset that is undamaged)

TABLE II
RECOGNITION RATES PER SUBSET OF MolVec

| Subset | Base | _20 | _40 | _60 | _80 | _99 | _5 | _10 | _15 | _20 | _25 | _30 |
|---|---|---|---|---|---|---|---|---|---|---|---|---|
| Blend | 89.1% | 89.1% | 88.4% | 89.1% | 89.1% | - | - | - | - | - | - | - |
| Compress | 89.1% | 86.8% | 86.8% | 89.1% | 84.5% | 43.4% | - | - | - | - | - | - |
| Convert | 89.1% | - | - | - | - | - | 88.4% | 81.4% | 48.1% | 31.0% | 43.4% | - |
| Distort | 89.1% | 74.4% | 75.2% | 70.5% | 62.0% | - | - | - | - | - | - | 62.0% |

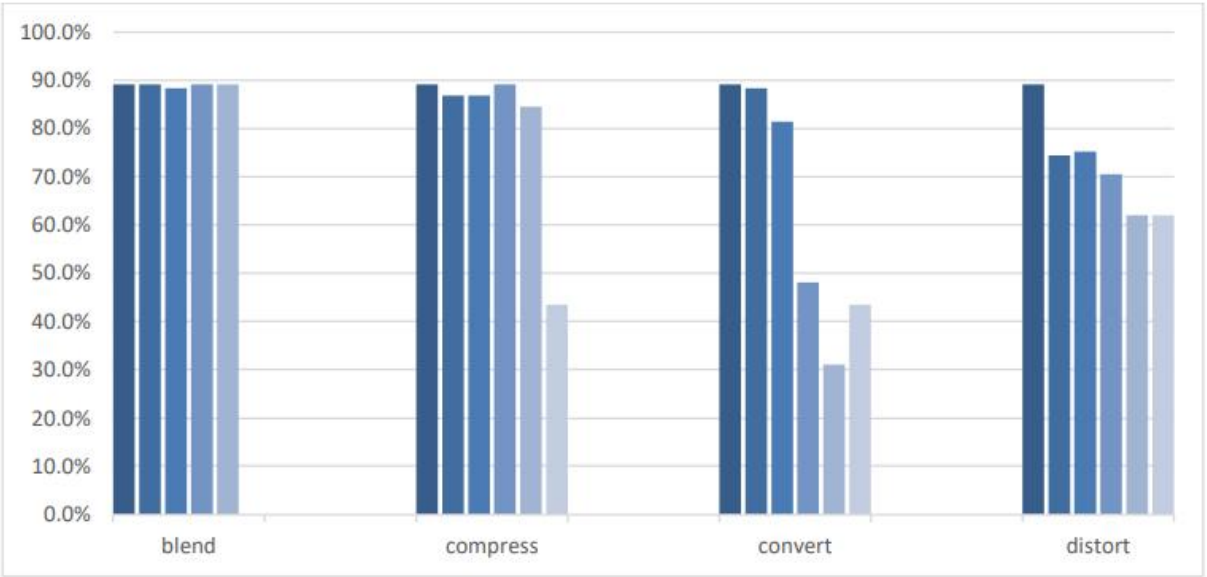

Fig. 11. Test results for MolVec subsets impacted by a particular kind of damage (the first column shows a subset that is undamaged)

The recognition performance of Decimer exhibited extreme sensitivity to the presence of a black overlay and noise as shown in Table III and Fig. 12, with both types of damage resulting in recognition success rates below 30%. In more severely damaged subsets, the recognition rate dropped to under 2%. This behavior can be attributed to Decimer's interpretation of the black overlay and noise as carbon atoms, leading to the generation of excessively long SMILES strings primarily composed of carbon atoms. Compression damage only began to notably impact Decimer's performance at the highest compression subset. However, Decimer demonstrated a notably high resilience to distorted structures, maintaining a relatively strong recognition ability under such conditions.

[19] https://www.rdkit.org/

The impact of a black overlay on MolScribe was inconsistent as shown in Table IV and Fig. 13, though overall negative, with recognition success dropping by approximately 50%. The highest percentage overlay resulted in a complete failure of recognition. MolScribe demonstrated a particularly strong ability to handle compressed images, with only a noticeable reduction in recognition at the 99% compression level. While MolScribe was able to tolerate low levels of noise reasonably well, higher noise levels caused recognition rates to fall below 20%. In terms of distortion, the effect was minimal at lower levels but became more pronounced as the distortion severity increased.

TABLE III
RECOGNITION RATES PER SUBSET OF DECIMER

| Subset | Base | _20 | _40 | _60 | _80 | _99 | _5 | _10 | _15 | _20 | _25 | _30 |
|---|---|---|---|---|---|---|---|---|---|---|---|---|
| Blend | 82.2% | 23.3% | 26.4% | 0.8% | 0.0% | - | - | - | - | - | - | - |
| Compress | 82.2% | 83.7% | 81.4% | 75.2% | 71.3% | 44.2% | - | - | - | - | - | - |
| Convert | 82.2% | - | - | - | - | - | 15.5% | 4.7% | 1.6% | 1.6% | 0.8% | - |
| Distort | 82.2% | 79.8% | 79.1% | 79.1% | 77.5% | - | - | - | - | - | - | 73.6% |

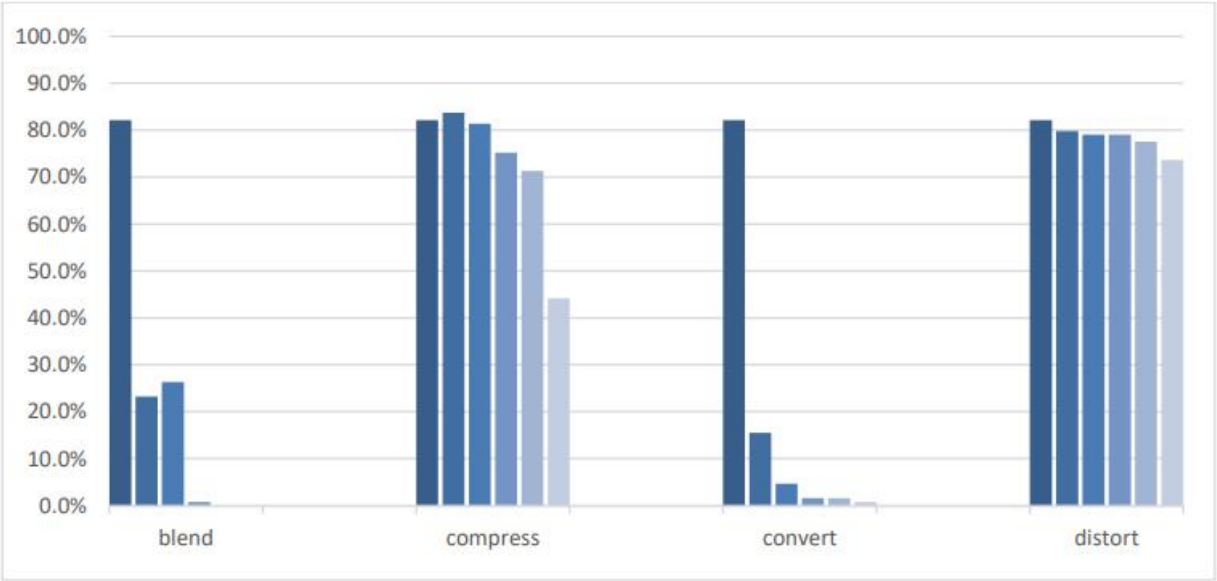

Fig. 12. Test results for Decimer subsets impacted by a particular kind of damage (the first column shows a subset that is undamaged)

TABLE IV
RECOGNITION RATES PER SUBSET OF MolScribe

| Subset | Base | _20 | _40 | _60 | _80 | _99 | _5 | _10 | _15 | _20 | _25 | _30 |
|---|---|---|---|---|---|---|---|---|---|---|---|---|
| Blend | 94.6% | 48.1% | 63.6% | 46.5% | 10.1% | - | - | - | - | - | - | - |
| Compress | 94.6% | 94.6% | 93.0% | 94.6% | 94.6% | 55.8% | - | - | - | - | - | - |
| Convert | 94.6% | - | - | - | - | - | 89.9% | 49.6% | 20.9% | 7.0% | 3.1% | - |
| Distort | 94.6% | 93.0% | 93.0% | 87.6% | 77.5% | - | - | - | - | - | - | 72.9% |

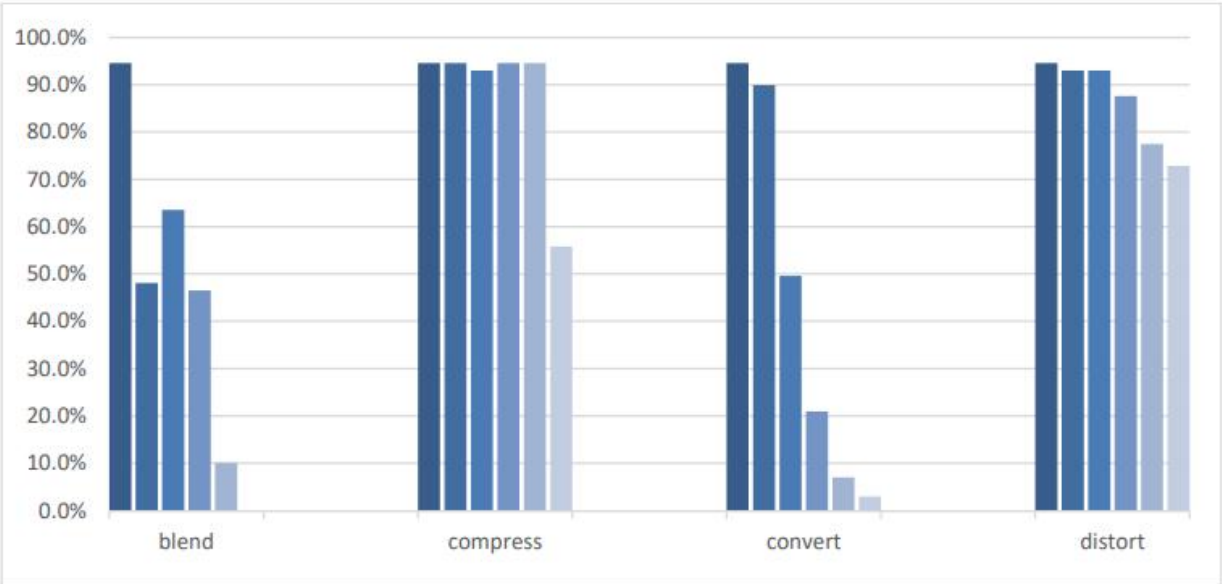

Fig. 13. Test results for MolScribe subsets impacted by a particular kind of damage (the first column shows a subset that is undamaged)

Fig. 14 presents the recognition rates, while Fig. 15 illustrates the percentage of performance decline resulting from the specified types and degrees of damage. Among the tested OCSR tools, MolScribe and MolVec achieved the highest recognition success rate on the undamaged subset, with Imago performing approximately 10% worse than the others. Notably, MolVec was the only tool unaffected by the black overlay, whereas Decimer exhibited particular sensitivity to this form of damage. Decimer also demonstrated higher vulnerability to noise compared to the other tools, while MolVec exhibited the greatest resistance to noise, though not complete immunity. All tools, except Imago, maintained a relatively high accuracy with compression rates up to 80%, although Decimer showed a noticeable decline in recognition at 60% and 80% compression rates. A significant reduction in accuracy, approximately 50%, was observed across all tools at the 99% compression rate subset. ML-based tools were more adept at handling distortion, with Decimer experiencing only a 10% reduction at the highest distortion level. In contrast, rule-based tools exhibited a recognition rate reduction in the range of 15-30%, with MolVec being the most vulnerable to distortion damage. While execution speed was not explicitly measured, the time taken by each tool to process the entire dataset differed significantly enough to warrant mention on a relative scale. It is important to note that execution time is influenced by numerous factors that may vary depending on the testing environment. MolVec and Imago processed the entire dataset within a matter of minutes, with MolVec demonstrating a noticeably faster processing time than Imago. In contrast, both Decimer and MolScribe required several hours to complete the dataset processing, with MolScribe being approximately six times slower than Decimer.

| | base | blend_20 | blend_40 | blend_60 | blend_80 |
|---|---|---|---|---|---|
| decimer | 82.2% | 23.3% | 26.4% | 0.8% | 0.0% |
| molscribe | 94.6% | 48.1% | 63.6% | 46.5% | 10.1% |
| imago | 73.6% | 34.9% | 33.3% | 32.6% | 29.5% |
| molvec | 89.1% | 89.1% | 88.4% | 89.1% | 89.1% |
| | convert_5 | convert_10 | convert_15 | convert_20 | convert_25 |
| decimer | 15.5% | 4.7% | 1.6% | 1.6% | 0.8% |
| molscribe | 89.9% | 49.6% | 20.9% | 7.0% | 3.1% |
| imago | 34.9% | 38.0% | 15.5% | 6.2% | 9.3% |
| molvec | 88.4% | 81.4% | 48.1% | 31.0% | 43.4% |
| | compress_20 | compress_40 | compress_60 | compress_80 | compress_99 |
| decimer | 83.7% | 81.4% | 75.2% | 71.3% | 44.2% |
| molscribe | 94.6% | 93.0% | 94.6% | 94.6% | 55.8% |
| imago | 37.2% | 37.2% | 36.4% | 34.9% | 20.2% |
| molvec | 86.8% | 86.8% | 89.1% | 84.5% | 43.4% |
| | distort_10 | distort_20 | distort_30 | distort_40 | distort_50 |
| decimer | 79.8% | 79.1% | 79.1% | 77.5% | 73.6% |
| molscribe | 93.0% | 93.0% | 87.6% | 77.5% | 72.9% |
| imago | 62.8% | 64.3% | 59.7% | 55.8% | 54.3% |
| molvec | 74.4% | 75.2% | 70.5% | 62.0% | 62.0% |

Fig. 14. Comparison of each evaluated tool's recognition rates by subset

| | base | blend_20 | blend_40 | blend_60 | blend_80 |
|---|---|---|---|---|---|
| decimer | 0.0% | 71.7% | 67.9% | 99.1% | 100.0% |
| molscribe | 0.0% | 49.2% | 32.8% | 50.8% | 89.3% |
| imago | 0.0% | 52.6% | 54.7% | 55.8% | 60.0% |
| molvec | 0.0% | 0.0% | 0.9% | 0.0% | 0.0% |
| | convert_5 | convert_10 | convert_15 | convert_20 | convert_25 |
| decimer | 81.1% | 94.3% | 98.1% | 98.1% | 99.1% |
| molscribe | 4.9% | 47.5% | 77.9% | 92.6% | 96.7% |
| imago | 52.6% | 48.4% | 78.9% | 91.6% | 87.4% |
| molvec | 0.9% | 8.7% | 46.1% | 65.2% | 51.3% |
| | compress_20 | compress_40 | compress_60 | compress_80 | compress_99 |
| decimer | -1.9% | 0.9% | 8.5% | 13.2% | 46.2% |
| molscribe | 0.0% | 1.6% | 0.0% | 0.0% | 41.0% |
| imago | 49.5% | 49.5% | 50.5% | 52.6% | 72.6% |
| molvec | 2.6% | 2.6% | 0.0% | 5.2% | 51.3% |
| | distort_10 | distort_20 | distort_30 | distort_40 | distort_50 |
| decimer | 2.8% | 3.8% | 3.8% | 5.7% | 10.4% |
| molscribe | 1.6% | 1.6% | 7.4% | 18.0% | 23.0% |
| imago | 14.7% | 12.6% | 18.9% | 24.2% | 26.3% |
| molvec | 16.5% | 15.7% | 20.9% | 30.4% | 30.4% |

Fig. 15. When comparing the percentage decrease in the recognition rate of damaged subsets, 0% indicates that the subset was recognized equally well as the undamaged, and 50% indicates that the recognition success rate of the damaged subset was half that of the undamaged

## VI. Conclusion and Future Works

Four distinct OCSR tools—two rule-based and two ML based—were installed and tested across multiple subsets featuring various levels and types of damage. The tools' performance was evaluated based on recognition success rates for both undamaged and damaged subsets, with particular attention given to the impact of different damage types on recognition success. Among the tested tools, MolScribe and MolVec distinguished themselves for different reasons. MolScribe demonstrated exceptional performance and robustness in handling compression and distortion damage, while MolVec stood out for its combination of reliable performance, high speed, and unique resilience to background blending and noise. Decimer exhibited the least sensitivity to distortion damage. Imago, though the easiest to install and use, was outperformed by the other tools across almost all subsets. With the exception of extreme compression, high levels of noise damage, and, to some extent, distortion, all other tested damage types were effectively handled without performance degradation by at least one of the tools. This suggests that incorporating specific strengths from different tools could yield better overall performance. For instance, integrating MolVec's resilience to background blending and noise into MolScribe could enhance its ability to handle such damage. Further testing could explore the inclusion of a binarization step in the process. Proper binarization could eliminate the impact of a blending black background, a common issue in scanned documents. This could address a potential limitation of many contemporary OCSR tools, which may lose recognition accuracy due to imperfect or missing binarization steps. An interesting direction for future work could involve combining two rule-based OCSR tools with one ML based tool (e.g., MolVec + OSRA + MolScribe). In this approach, if MolVec and OSRA outputs match, they would be accepted as the final output; if not, MolScribe could be used as the fallback. This hybrid model could provide a balance between execution speed and performance, leveraging the strengths of both rule-based and ML based methods.

## VII. Declarations

*A.* ***Funding:*** No funds, grants, or other support was received.

*B.* ***Conflict of Interest:*** The authors declare that they have no known competing for financial interests or personal relationships that could have appeared to influence the work reported in this paper.

*C.* ***Data Availability:*** Data will be made on reasonable request.

***D. Code Availability:*** Code will be made on reasonable request.